\documentclass[conference]{IEEEtran}
\usepackage[utf8]{inputenc}
\usepackage{graphicx}
\usepackage{xcolor}
\usepackage{url}
\usepackage{mdframed}
\usepackage{subfig}
\usepackage{tabularx}
\usepackage{multirow}
\usepackage{colortbl}
\usepackage{booktabs} 

\usepackage{amssymb}
\usepackage{amsmath}

\usepackage{mdframed}

\definecolor{maroon}{cmyk}{0,0.87,0.68,0.32}

\begin{document}
\title{Classifying Multilingual User Feedback using Traditional Machine Learning and Deep Learning}

\author{\IEEEauthorblockN{Christoph Stanik, Marlo Haering and Walid Maalej
\IEEEauthorblockA{University of Hamburg\\
Hamburg, Germany\\
\{stanik, haering, maalej\}@informatik.uni-hamburg.de}}}

\maketitle

\begin{abstract}
With the rise of social media like Twitter and of software distribution platforms like app stores, users got various ways to express their opinion about software products. 
Popular software vendors get user feedback thousandfold per day.
Research has shown that such feedback contains valuable information for software development teams such as problem reports or feature and support inquires. 
Since the manual analysis of user feedback is cumbersome and hard to manage many researchers and tool vendors suggested to use automated analyses based on traditional supervised machine learning approaches.
In this work, we compare the results of traditional machine learning and deep learning in classifying user feedback in English and Italian into problem reports, inquiries, and irrelevant.
Our results show that using traditional machine learning, we can still achieve comparable results to deep learning, although we collected thousands of labels.
\end{abstract}
\begin{IEEEkeywords}
    Data-Driven Requirements, 
    Data Mining, 
    Social Media Analytics, 
    Machine Learning, 
    Deep Learning
\end{IEEEkeywords}
\IEEEpeerreviewmaketitle

\section{Introduction} \label{sec:introduction}
\textbf{Motivation.}
Research has shown the importance of extracting requirements related information from user feedback to improve software products and user satisfaction \cite{palomba2015user}. 
As user feedback on social media or app stores can come thousandfold daily, a manual analysis of that feedback is cumbersome \cite{Pagano:App:2013}.
However, analyzing this feedback brings opportunities to understand user opinions better because it contains valuable information like problems users encounter or features they miss \cite{Pagano:App:2013, Guzman:Twitter:2016}.
Researchers have applied supervised machine learning to filter noisy, irrelevant feedback and to extract requirements related information \cite{Maalej:REJ:2016, Guzman:Twitter:2017}.
Most related works rely on traditional machine learning approaches, which require domain experts to represent the data with hand-crafted features.
In contrast, end-to-end deep learning approaches automatically learn high-level feature representations from raw data without domain knowledge, achieving remarkable results in different classification tasks \cite{goodfellowDeepLearning2016, song2015end, youngRecentTrendsDeep2017}.
\\
\textbf{Objective.}
In this work, we aim at understanding if and to what extent deep learning can improve state-of-the-art results for classifying user feedback into problem reports, inquiries, and irrelevant.
We focus on these three categories because practitioners seek for automated solutions to filter noisy feedback (irrelevant), to identify and fix bugs (problem reports), and to find feature requests as inspiration for future releases (inquiries) \cite{Maalej:REJ:2016}.
We consider all user feedback as problem reports, that state a concrete problem related to a software product or service (e.g., ``Since the last update the app crashes upon start'').
We define inquires as user feedback that asks for either new functionality, an improvement, or requests information for support (e.g., ``It would be great if I could invite multiple friends at once'').
We consider user feedback as irrelevant if it does not belong to problem reports or inquires (e.g., ``I love this app'').

To fulfill our objective, we employ supervised machine learning fed with crowd-sourced annotations of 10,000 English and 15,000 Italian tweets from telecommunication Twitter support accounts, and 6,000 annotations of English app reviews.
We apply best practices for both machine learning approaches (traditional and deep learning) and report on a benchmark.
\\
\textbf{Preliminary results.}
Our preliminary results show that, within our setting, traditional machine learning can achieve comparable results to deep learning.
One possible explanation is that domain experts' knowledge in traditional machine learning brings considerable performance improvements using simple but powerful features, including specific keywords.
In general, the classification of irrelevant user feedback achieves the best results meaning that practitioners could use our reported models to filter noisy feedback.
\\
\textbf{Contribution.} 
The contribution of this paper is threefold. First, we give insights on how traditional machine learning compares to deep learning on classifying feedback by describing both approaches and by performing a large series of experiments. Second, we provide a replication package containing the scripts and experiment setups. Third, we report the configurations of top-performing machine learning models.
\\
\textbf{Structure.}
In Section \ref{sec:methodology}, we introduce the methodology of this paper by detailing our research questions, design, and data.
Section \ref{sec:pipeline} describes the pipeline and the setup for both machine learning approaches.
Section \ref{sec:results} reports on our classification benchmark showing the accuracy and the configuration of the top-performing  models. 
Then, Section \ref{sec:discussion} discusses the implications of the results and possible application fields, as well as the threats to validity.
Section \ref{sec:related_work} summarizes the related work while Section \ref{sec:conclusion} concludes the paper.
\section{Methodology} \label{sec:methodology}
We discuss the research questions, as well as our study design, and the data our analysis rely on.
\subsection{Research Question} \label{ssec:research_questions}
The goal of this work is to identify the top-performing model to classify user feedback (tweets and app reviews) into problem reports, inquires, and irrelevant by comparing the traditional machine learning approach with deep learning.
We, therefore, state the following research questions:
\begin{itemize}
    \item \textbf{RQ1.} To what extent can we extract problem reports, inquires, and irrelevant information from user feedback using traditional machine learning?
    \item \textbf{RQ2.} To what extent can we extract problem reports, inquires, and irrelevant  information from user feedback using deep learning?
    \item \textbf{RQ3.} How do the results of the traditional machine learning approach and the deep learning approach compare and what can we learn from it?
\end{itemize}

\subsection{Study Design} \label{ssec:study_design}
\begin{figure}[ht]
    \centering
    \includegraphics[width=\columnwidth]{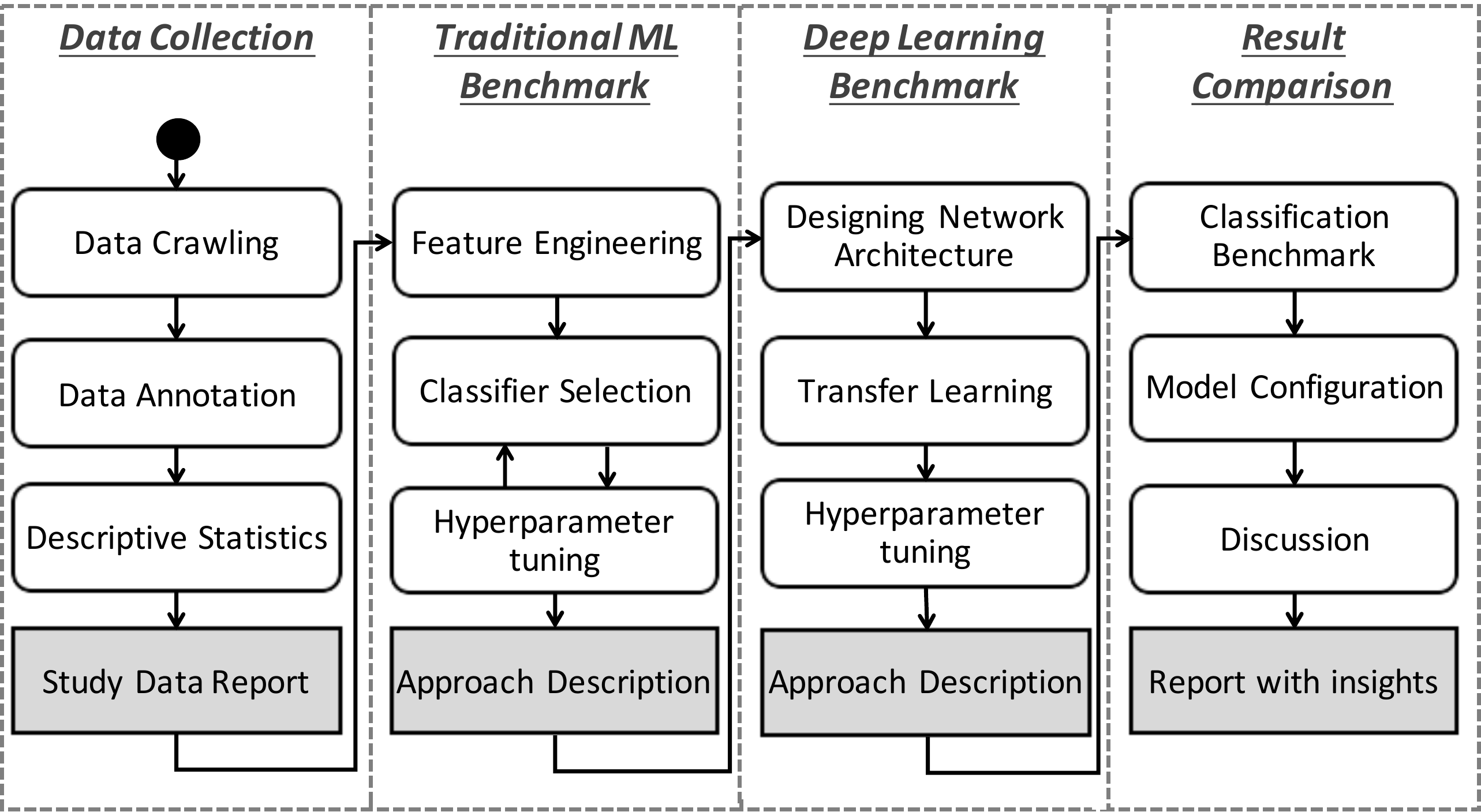}
    \caption{Overview of the study design.}
    \label{fig:study_design}
\end{figure}
Figure \ref{fig:study_design} shows the overall study design. Each white box within the four columns describes a certain step that we performed while the grey boxes show the result that each column produced.
The first part of the paper is about the \textit{Study Data}, which we describe later in this section.
In the second part, \textit{Traditional Approach}, we perform traditional machine learning engineering, including feature engineering and hyper-parameter tuning.
In the part \textit{Deep Learning Approach}, we design a convolutional neural network architecture, apply transfer learning for the embedding layer, and finally evaluate the fine-tuned models.
In the fourth part \textit{Result Comparison}, we report on the results of our classification experiments (benchmark) comparing the traditional and the deep learning approaches.

\subsection{Study Data} \label{ssec:study_data}
\begin{table}[]
    \centering
    \footnotesize
    \caption{Overview of the study data.}
    \label{tab:study_data}
    \begin{tabular}{l|c|rr}
    \toprule
                       & App Reviews & \multicolumn{2}{c}{Tweets} \\
                       & \multicolumn{1}{c|}{English}     & English      & Italian      \\ \midrule
    n\_problem\_report & 1.437       & 2.933        & 3.414     \\
    n\_inquiry         & 1.100       & 1.405        & 2.594     \\
    n\_irrelevant      & 3.869       & 6.026        & 9.794     \\ \midrule
    \rowcolor{gray!20}TOTAL              & 6.406       & 10.364       & 15.802    \\ \bottomrule
\end{tabular}
\end{table}

We collected about 5,000,000 English and 1,300,000 Italian Tweets addressing Twitter support accounts of telecommunication companies.
From that corpus, we randomly sampled $\sim$10,000 English tweets and $\sim$15,000 Italian tweets that were composed by users.
As the annotation of so many tweets is very time-consuming, we created coding tasks on the crowd-annotation platform \emph{figure eight}\footnote{\url{https://www.figure-eight.com/}}.
Before starting the crowd-annotation, we first wrote a coding guide to describe our understanding of problem reports, inquiries, and irrelevant tweets with the help of the innovation center of a big Italian telecommunication company.
Second, we run a pilot study to test the quality of the coding guide and the annotations received.
Both coding guides were either written or proof-read by at least two native speakers, and we required that the annotators are natives in the language.
Each tweet can belong to exactly one of the before mentioned classes and is annotated by at least two persons, three in case of a disagreement.
As for the annotated app reviews, we rely on the data and annotations of Maalej et al. \cite{Maalej:REJ:2016}.
Table \ref{tab:study_data} summarizes the annotated data for both languages.

\textbf{Replication package}. To encourage replicability, we uploaded all scripts, benchmark results, and provide the annotated dataset upon request\footnote{https://mast.informatik.uni-hamburg.de/replication-packages/}.
\section{Machine Learning Pipelines} \label{sec:pipeline}
We describe how we performed the machine learning approaches and explain certain decisions such as for the selected features.
To ensure a fair comparison between the traditional and the deep learning approach, we used not only the same datasets but also the same train and test sets.

\subsection{Traditional Machine Learning} \label{ssec:traditional}
\subsubsection{Preprocessing}
We preprocessed the data in three steps to reduce ambiguity.
Step 1 turns the text into lower case; this reduces ambiguity by normalizing, e.g., ``Feature'', ``FEATURE'', and ``feature'' by transforming it into the same representation ``feature''.
Step 2 introduces masks to certain keywords. 
For example, whenever an account is addressed using the ``@'' symbol, the account name will be masked as ``account''. 
We masked account names, links, and hashtags.
Step 3 applies lemmatization, which normalizes the words to their root form. 
For example, words such as ``see'', ``saw'', ``seen', and ``seeing'' become the word ``see''.

\subsubsection{Feature Engineering}
\begin{table}[]
    \centering
    \footnotesize
    \caption{Extracted features before scaling. If not further specified, the number of features applies to all data sets.}
    \label{tab:features}
    \begin{tabular}{lll}
    \toprule
    Feature Group         & Value Boundaries & Number of Features \\ \midrule \rowcolor{gray!20}
    n\_words        & $\mathbb{N}$    & 1 \\
    n\_stopwords    & $\mathbb{N}$    & 1 \\\rowcolor{gray!20} 
    sentiment$_{neg}$ & $\{x \in \mathbb{Z} \, | \,-5 \le x \le -1\}$ & 1 \\\rowcolor{gray!20} 
    sentiment$_{pos}$ & $\{x \in \mathbb{N} \, | \,1 \le x \le 5\}$     & 1 \\
    keywords        & $\{0,1\}$                 & 37 (IT), 60 (EN) \\\rowcolor{gray!20}
    POS tags        & $\mathbb{N}$    & 18 (IT), 16 (EN) \\
    tense           & $\mathbb{N}$    & 4 (IT), 2 (EN)   \\ \rowcolor{gray!20}
    tf-idf          & $\{x \in \mathbb{R} \, | \,0 \le x \le 1\}$ & 665 (app reviews, EN)        \\\rowcolor{gray!20}
    && 899 (tweets, EN)\\\rowcolor{gray!20}
    && 938 (tweets IT) \\
    fastText        & $\{x \in \mathbb{R} \, | \,0 \le x \le 1\}$ & 300 \\\rowcolor{gray!20}\midrule
    TOTAL && 1.047 (app reviews, EN)\\\rowcolor{gray!20}
    && 1.281 (tweets, EN)\\\rowcolor{gray!20}
    && 1.301 (tweets IT)\\ \bottomrule
    \end{tabular}
\end{table}
Feature engineering describes the process of utilizing domain knowledge to find a meaningful data representation for machine learning models.
In NLP it encompasses steps such as extracting features from text, as well as selection and optimization.
Table \ref{tab:features} summarizes the groups of features, their representation, as well as the number of features we extracted for that feature group.
For instance, the table shows that the feature group ``keywords'' consists of 37 keywords for the Italian language, each of them being 1 if that keyword exists or 0 if not.

We extracted the \emph{length (n words)} of the written user feedback as Pagano and Maalej \cite{Pagano:App:2013} found that most irrelevant reviews are rather short.
One example for such a category is \emph{rating}, which does not contain valuable information for developers as most of the time, such reviews are only praise (e.g., ``I love this app.'').
Excluding or including \emph{stop words}, in particular in the preprocessing phase is highly discussed in research.
We found papers that reported excluding stop words as an essential step (e.g., \cite{Guzman:Sentiment:2014}), papers that leveraged the inclusion of certain stop words (e.g., \cite{Johann:2017:app}), and others that tested both (e.g.,  \cite{Maalej:REJ:2016}).
However, the decision for exclusion and inclusion depends on the use case.
We decided to use them as a feature by counting their occurrence in each document.

Further, we extracted the \emph{sentiment} of the user feedback using the sentistrength library \cite{thelwall2010sentiment}.
We provide the full user feedback (e.g., a tweet) as the input for the library.
The library then returns two integer values, one ranging from -5 to -1 indicating on how negative the feedback is, the other ranging from +1 to +5 indicating how positive the feedback is.
The sentiment can be an important feature as users might write problem reports in a neutral to negative tone while inquiries tend to be rather neutral to positive \cite{Guzman:Sentiment:2014, Pagano:App:2013, Maalej:REJ:2016}. 
\emph{Keywords} have proven to be useful features for text classification \cite{villarroel:release:2016, Maalej:REJ:2016, haringWhoAddressedThis2018} as their extraction allows input of domain experts' knowledge.
However, keywords are prone to overfit for a single domain and therefore might not be generalizable.
In this work, we use the same set of keywords for the English app reviews and tweets.
We extracted our set of keywords by 1) looking into related work \cite{Iacob:Review:2013, villarroel:release:2016, Maalej:REJ:2016}, and 2) by manually analyzing 1,000 documents from the training set of all three datasets following the approach from Iacob and Harrison \cite{Iacob:Review:2013}.
Kurtanovi\'c and Maalej \cite{Kurtanovic:Mining:2017, Kurtanovic:Req:2017} successfully used the counts of \emph{Part-of-speech (POS) tags} for classification approaches in requirements engineering. Therefore we also included them in our experiments.

Maalej et al.~\cite{Maalej:REJ:2016} successfully utilized the \emph{tenses} of sentences.
This feature might be useful for the classification as users write problem reports often in the past or present tense, e.g., ``I updated the app yesterday. Since then it crashes.'' and inquiries (i.e., feature requests) in the present and future tense, e.g., ``I hope that you will add more background colors''.
When extracting the tense using spaCy\footnote{\url{https://spacy.io/}} the Italian language model supported four tenses while for the English language we had to deduce the tense by extracting the part-of-speech tags.
\emph{Tf-idf (term  frequency-inverse  document  frequency)} \cite{Sparck:Tfidf:1972} is a frequently used technique to represent text in a vector space.
It increases proportionally to the occurrence of a term in a document but is offset by the frequency of the term in the whole corpus. 
Tf-idf combines term frequencies with the inverse document frequency to calculate the term weight in the document.

\emph{FastText} \cite{Joulin:Fasttext:2016} is an unsupervised approach to learn high-dimensional vector representations for words from a large training corpus.
The vectors of words that occur in a similar context are close in this space.
Although the fastText library provides pre-trained models for several languages, we train our own domain-specific models based on 5,000,000 English app reviews, 1,300,000 Italian tweets, and 5,000,000 Italian tweets.
We represent each document as the average vector of all word vectors of the document, which is also a 300-dimensional vector.
We chose fastText for our word embedding models as it composes a word embedding from subword embeddings. In contrast, word2vec \cite{Mikolov:W2V:2013} learns embeddings for whole words.
Thereby, our model is able to 1) recognize words that were not in the training corpus and 2) capture spelling mistakes, which is a typical phenomenon in user feedback.

\begin{figure*}[tbh]
    \centering
    \includegraphics[width=\textwidth]{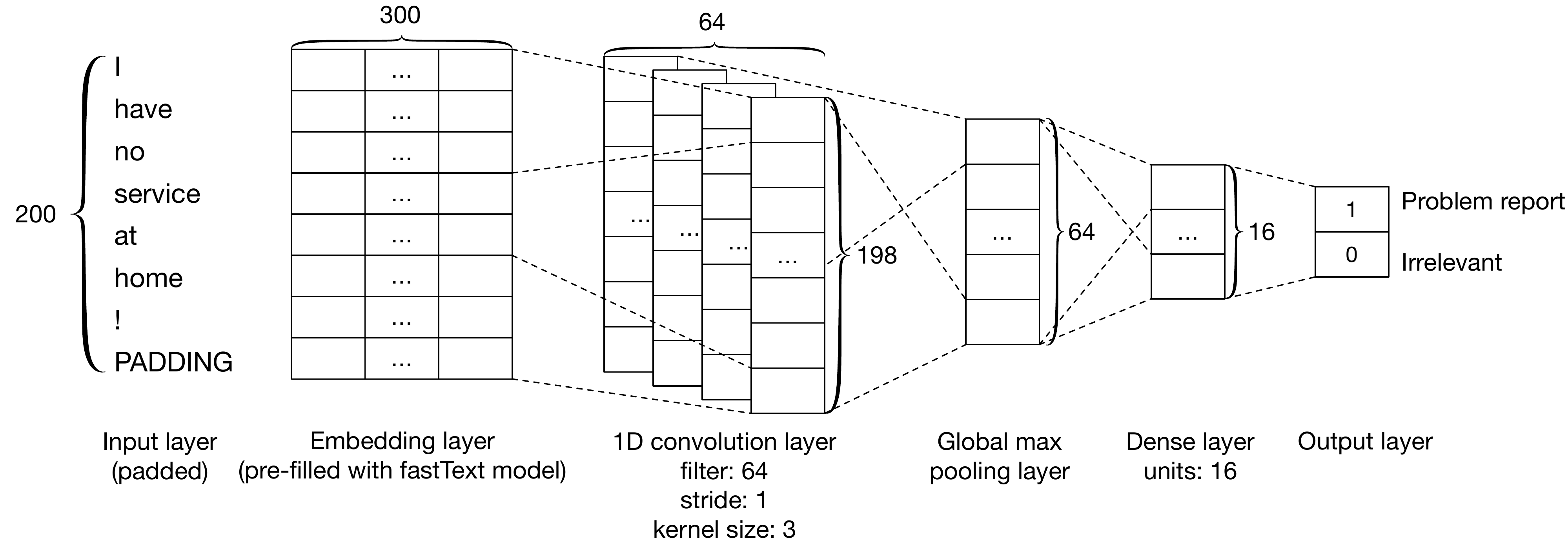}
    \caption{Neural network architecture for the classification.}
    \label{fig:cnn_model}
\end{figure*}
\subsubsection{Experiment Configuration}
For the experiment setup, we tried to find the most accurate  machine learning model by varying five dimensions (no particular order).
In the \emph{first dimension} we target to find the best-performing features of Table \ref{tab:features} by testing different combinations.
In total, we tested 30 different feature combinations such as ``sentiment + fastText'' and ``n\_words + keywords + POS tags + tf-idf''.

The \emph{second dimension} is testing the performance of (not) applying feature scaling.
Tf-idf vectors, for example, are represented by float numbers between 0 and 1, while the number of words can be any number greater than 0.
This could lead to two issues: 1) the machine learning algorithm might give a higher weight to features with a high number meaning that the features are not treated equally. 2) the machine learning model could perform worse if features are not scaled. 

In the \emph{third dimension}, we perform Grid Search \cite{Bergstra:Param:2011} for hyper-parameter tuning. In contrast to Random Search, which samples hyper-parameter combinations for a fixed number of settings \cite{Bergstra:Param:2012}, Grid Search exhaustively combines hyperparameters of a defined grid. 
For each hyper-parameter combination in the Grid Search, we perform 5-fold cross-validation of the training set. 
We optimize the hyperparameters for the f1 metric to treat precision and recall as equally important. 

The \emph{fourth dimension} checks whether sampling (balancing) the training data improves the overall performance of the classifiers. 
For unbalanced data the machine learning algorithm might tend to categorize a document as part of the majority class as this is the most likely option.
In this work we test both, keeping the original distribution of documents per class and applying random under-sampling on the majority class to create a balanced training set. 

Finally, the \emph{fifth dimension} is about testing different machine learning algorithms. 
Similar to our reasoning for the feature selection, we tested the following algorithms frequently used in related work: Decision Tree, Random Forest, Naive Bayes, and Support Vector Machine \cite{Maalej:REJ:2016, Guzman:Twitter:2017, Williams:Twitter:2017}. 
As for the classification, we follow the insights from Maalej et al.~\cite{Maalej:REJ:2016} and employ binary classification (one for each: problem report, inquiry, and irrelevant) instead of multiclass classification. 

\subsection{Deep Learning} \label{ssec:deep_learning}
\subsubsection{Deep Learning}
Traditional classification approaches require a data representation based on hand-crafted features, which domain experts deem useful criteria for the classification problem at hand. 
In contrast, neural networks, which are used in deep learning approaches, use the raw text as an input, and learn high-level feature representations automatically \cite{goodfellowDeepLearning2016}. 
In previous work, researchers applied them in diverse applications with remarkable results to different classification tasks, including object detection in images, machine translation, sentiment analysis, and text classification tasks \cite{collobertNaturalLanguageProcessing2011}. However, neural networks are not a silver bullet, and they have also achieved only  moderate results in the domain of software engineering  \cite{guzmanEnsembleMethodsApp2015, Fakhoury:SANER:2018, Fu:FSE:2017}.

\begin{table*}[]
    \centering
    \footnotesize
    \caption{Classification benchmark for the traditional machine learning approach (Trad.) and the deep learning approach (DL). The best f1 score per classification problem and dataset is marked in bold font.}
    \label{tab:benchmark}
    \begin{tabular}{ll|cccc|cccc|cccc}
    \toprule
     &  & \multicolumn{4}{c}{app review EN} & \multicolumn{4}{c}{tweet EN} & \multicolumn{4}{c}{tweet IT} \\
     
                                                          & & p   & r   & f1  & auc & p & r & f1 & auc  & p   & r   & f1  & auc \\ \midrule
\multirow{3}{*}{\rotatebox[origin=c]{90}{\textit{Trad.}}} & problem report & .83 & .75 & \textbf{.79} & .85 & .46 & .82 & \textbf{.59} & .72 & .51 & .88 & \textbf{.65} & .83 \\ 
     & inquiry &.68 & .76 & .72 & .85 & .32 & .70 & \textbf{.43} & .73 & .47 & .82 & \textbf{.60} & .82\\
     & irrelevant & .88 & .89 & \textbf{.89} & .86 & .73 & .75 & \textbf{.74} & .69 & .78 & .89 & \textbf{.83} & .73\\ \midrule
    \multirow{3}{*}{\rotatebox[origin=c]{90}{\textit{DL}}} & problem report & .46 & .60 & .52 & .82 & .51 & .42 & .46 & .74 & .62 & .57 & .59 & .84 \\
     & inquiry & .69 & .79 & \textbf{.74} & .94 & .40 & .40 & .40 & .75 & .51 & .57 & .54 & .83\\
     & irrelevant & .78 & .93 & .85 & .90 & .74 & .70 & .72 & .75 & .85 & .77 & .81 & .86\\ \bottomrule
    \end{tabular}
\end{table*}

\subsubsection{Convolutional Neural Networks}
Although convolutional neural networks (CNNs) have mainly been used for image classification tasks, researchers also applied them successfully to natural language processing problems \cite{kimConvolutionalNeuralNetworks2014,lopezDeepLearningApplied2017}. 
In most cases, deep learning approaches require a large amount of training data to outperform traditional approaches. 
Figure \ref{fig:cnn_model} shows the architecture of the neural network that we used for the experiments in this study. Häring et al.~\cite{haringWhoAddressedThis2018} used this model to identify user comments on online news sites that address either the media house, the journalist, or the forum moderator.
They achieved promising results that partly outperformed a traditional machine learning approach. 
The input layer requires a fixed size for the text inputs. 
We choose the size 200, which we found appropriate for both the app review and the Twitter dataset as tweets are generally shorter and we identified less than 20 app reviews that exceed 200 words. 
We cut the part, which is longer than 200 words and pad shorter input texts, so they reach the required length. 
After the input layer our network consists of an embedding layer, a 1D convolution layer, a 1D global max pooling layer, a dense layer, and a concluding output layer with a softmax activation. 
For the previous layers, we used the tanh activation function. During training, we froze the weights of the embedding layer, whereby ~15,000 trainable parameters remain.

\subsubsection{Transfer Learning}
Transfer learning is a method often applied to deep learning using models pre-trained on another task \cite{goodfellowDeepLearning2016}. 
In natural language processing, a common application of transfer learning is to reuse word embedding models, e.g. word2vec \cite{Mikolov:W2V:2013} or fastText \cite{Joulin:Fasttext:2016}, which were previously trained on a large corpus to pre-initialize the weights of an embedding layer. 
We applied transfer learning to pre-initialize our embedding layer with  three different pre-trained fastText models \cite{Joulin:Fasttext:2016}. During training, we froze the weights of the embedding layer.

\subsubsection{Hyperparameter Tuning}
The network architecture and the hyperparameter configuration can be a crucial factor for the performance of the neural network. 
Therefore we compared variations of both our CNN architecture as well as training parameters and evaluated the best-performing model on the test set. 
We performed a grid search and varied the number of filters and the kernel size of the 1D convolutional layer, the number of units for the dense layer, the number of epochs and the batch size for the training, and the number of units for the final dense layer. 
Due to the small size of our training set, we conducted a stratified 3-fold cross-validation on the training set for each hyperparameter configuration to acquire reliable results. 
Subsequently, we evaluated the model with the best-performing hyperparameter configuration on the test set. 
We trained the models with seven epochs and a batch size of 32. 
We used the Python library Keras \cite{chollet2015keras} for composing, training, and evaluating the models.

\section{Results} \label{sec:results}
\begin{table*}[]
    \centering
    \footnotesize
    \caption{Configuration of the best performing classification experiments for the traditional machine learning and the deep learning approaches. RF = Random Forest, DT = Decision Tree. CNN = Convolutional Neural Network.}
    \label{tab:configuration}
    \begin{tabular}{lll|l}
    \toprule
\multirow{18}{*}{\rotatebox[origin=c]{90}{\textit{Traditional Machine Learning}}} &
\multirow{6}{*}{app review EN}
    & \multirow{2}{*}{problem report} 
        & RF(max\_features:None, n\_estimators:500). \\
        &&&\textbf{features}:sentiment, tfidf, \textbf{sampling}:true, \textbf{scaling}:false \\ 
    && \multirow{2}{*}{inquiry}
        & DT(criterion:gini, max\_depth:1, min\_samples\_leaf:1, min\_samples\_split:4, splitter:random). \\
        &&&\textbf{features}:tfidf, keywords, \textbf{sampling}:false, \textbf{scaling}:false \\
    && \multirow{2}{*}{irrelevant}
        & DT(criterion:gini, max\_depth:8, min\_samples\_leaf:2, min\_samples\_split:4, splitter:random). \\ 
        &&&\textbf{features:}n\_words,n\_stopwords, n\_tense, n\_pos, keywords, tfidf, \textbf{sampling}:false, \textbf{scaling}:false \\ \cline{2-4}
& \multirow{6}{*}{tweet EN}
    & \multirow{2}{*}{problem report}
        & RF(max\_features:auto, n\_estimators:1000). \\
        &&&\textbf{features:}sentiment, tfidf, \textbf{sampling}:true, \textbf{scaling}:true \\ 
    && \multirow{2}{*}{inquiry}
        & DT(criterion:gini, max\_depth:1, min\_samples\_leaf:1, min\_samples\_split:2, splitter:best). \\
        &&&\textbf{features:}n\_words,n\_stopwords, n\_tense, n\_pos, keywords, tfidf, fastText, \textbf{sampling}:true, \textbf{scaling}:true \\
    && \multirow{2}{*}{irrelevant}
        & RF(max\_features:none, n\_estimators:1000). \\
        &&&\textbf{features:}n\_words,n\_stopwords, n\_tense, n\_pos, keywords, fastText, \textbf{sampling}:true, \textbf{scaling}:false\\ \cline{2-4}
& \multirow{6}{*}{tweet IT}
    & \multirow{2}{*}{problem report}
        & RF(max\_features:log2, n\_estimators:1000) \\
        &&& \textbf{features:}sentiment, n\_words,n\_stopwords, n\_tense, n\_pos, tfidf, \textbf{sampling}:true, \textbf{scaling}:true \\ 
    && \multirow{2}{*}{inquiry}
        & DT(criterion:entropy, max\_depth:8, min\_samples\_leaf:10, min\_samples\_split:6, splitter:random) \\
        &&& \textbf{features:}n\_words,n\_stopwords, n\_tense, n\_pos, keywords, \textbf{sampling}:true, \textbf{scaling}:false \\
    && \multirow{2}{*}{irrelevant}
        & DT(criterion:entropy, max\_depth:8, min\_samples\_leaf:8, min\_samples\_split:2, splitter:random) \\
        &&& \textbf{features:}sentiment, n\_words,n\_stopwords, n\_tense, n\_pos, tfidf, keywords, \textbf{sampling}:false, \textbf{scaling}:true\\ \midrule\midrule
\multirow{9}{*}{\rotatebox[origin=c]{90}{\textit{Deep Learning}}} &
\multirow{3}{*}{app review EN}
    & problem report 
        & CNN(dense\_number\_units:32, kernel\_size:3, number\_filters:16). \textbf{sampling}:true, \textbf{scaling}:true \\
    && inquiry 
        & CNN(dense\_number\_units:32, kernel\_size:5, number\_filters:16). \textbf{sampling}:true, \textbf{scaling}:true \\
    && irrelevant 
        & CNN(dense\_number\_units:32, kernel\_size:5, number\_filters:16). \textbf{sampling}:true, \textbf{scaling}:true \\ \cline{2-4}
& \multirow{3}{*}{tweet EN}
    & problem report 
        & CNN(dense\_number\_units:32, kernel\_size:5, number\_filters:16). \textbf{sampling}:true, \textbf{scaling}:true \\ 
    && inquiry 
        & CNN(dense\_number\_units:16, kernel\_size:5, number\_filters:16). \textbf{sampling}:true, \textbf{scaling}:true \\
    && irrelevant 
        & CNN(dense\_number\_units:32, kernel\_size:5, number\_filters:16). \textbf{sampling}:true, \textbf{scaling}:true \\ \cline{2-4}
& \multirow{3}{*}{tweet IT}
    & problem report 
        & CNN(dense\_number\_units:32, kernel\_size:5, number\_filters:16). \textbf{sampling}:true, \textbf{scaling}:true \\ 
    && inquiry 
        & CNN(dense\_number\_units:32, kernel\_size:5, number\_filters:16). \textbf{sampling}:true, \textbf{scaling}:true \\
    && irrelevant 
        & CNN(dense\_number\_units:32, kernel\_size:5, number\_filters:16). \textbf{sampling}:true, \textbf{scaling}:true \\ \bottomrule
    \end{tabular}
\end{table*}

In this section, we describe and discuss the results of the classification experiments. 
We first explain the evaluation metrics. 
Then, we report on the benchmark in Table \ref{tab:benchmark} showing the top accuracy. 
Finally, we explain the configuration of the models leading to the best results from Table \ref{tab:configuration}.

For this work, we report on the classification metrics \emph{precision}, \emph{recall}, and \emph{f1} as presented in related work \cite{Guzman:Sentiment:2014, villarroel:release:2016, Maalej:REJ:2016}.
For the calculation of these metrics we used sklearn's strictest parameter setting \emph{average=binary}, which is only reporting the result for classifying the true class. 
Additionally, we report on the Area Under the Curve \emph{AUC} value, which is considered a better metric when dealing with unbalanced data as it is independent of a certain threshold for binary classification problems.
In machine learning, Area Under the Receiver Operating Characteristics \emph{ROC AUC} is a metric frequently used to address class imbalance. Davis and Goadrich \cite{Davis:Metric:2006} argue that Precision-Recall AUC (PR AUC) is a more natural evaluation metric for that problem.
We optimized and selected the classification models based on f1, the harmonic mean of precision and recall. Thereby  either precision or recall can have a rather low value compared to the other.

Table \ref{tab:benchmark} shows the classification results of the best model for each of the three data sets and each classification problem.
For the English app reviews,  traditional machine learning  generally performs better than deep learning when considering the f1 score.
One reason for this difference might be, that for the app reviews, we have only about 6,000 annotated data points while for tweets we have about 10,000 for English tweets and about 15,000 for Italian tweets.
For the English tweets, both approaches perform quite similar.
While the f1 score seems to be lower for the deep learning approach, the AUC values are similar for both approaches.
The results for the Italian tweets show when optimizing towards f1, that deep learning reaches a higher precision, while the traditional approaches achieve a higher recall. The f1 score reveals again that both approaches perform similarly.
Based on our results, which are generated by a large series of experiments, we cannot say that for our setup, either of the approaches performs better.

\section{Discussion} \label{sec:discussion}
\subsection{Implications of the Results} \label{ssec:implication}
In this work, we classified user feedback for two languages from two different feedback channels.
We found that when considering the f1 score as a measure, traditional machine learning performs slightly better in most of the examined cases.
We expect that our approaches can also be applied to further feedback channels and languages, although some features are language-dependent and need to be updated.
For example, our deep learning model requires on top of a training set a pre-trained word embedding model for each  language such as the English and Italian fastText models used. 
Word embeddings capture the similarity between words depending on the domain and language. They are highly adaptable to language development by retraining the model regularly on current app reviews and tweets. 
It can capture the meaning of transitory terms like Twitter hashtags or emoticons. 
In traditional approaches, the language-dependent features are keywords, sentiment, POS tags, and the tf-idf vocabulary. This  requires more effort for creating models for multiple languages.
The rest remains language and domain-independent.

Traditional approaches often perform better on small training sets as domain experts implicitly incorporate significant information through hand-crafted features \cite{cholletDeepLearningPython2018}. 
We assume that for these experiments, the hand-crafted features derived from the domain experts lead to considerably better classification results. 
Deep neural networks derive high-level features automatically by utilizing large training samples. We presume that with more training data, a deeper neural network would outperform the traditional approach.

\subsection{Field of Application} \label{ssec:field_of_application}
Classifying user feedback is an ongoing field in research because of the high amount of feedback companies receive daily.
Pagano and Maalej \cite{Pagano:App:2013} show that, back in 2012, visible app vendors  receive, on average, 22 reviews per day in the app stores. 
Free apps receive a significantly higher amount of reviews ($\sim$37 reviews/day) compared to paid apps ($\sim$7 reviews/day). Popular apps such as Facebook receive about 4,000 reviews each day. 
When considering Twitter as a data source for user feedback for apps Guzman et al. \cite{Guzman:Twitter:2016} show that popular app development companies receive on average about 31,000 daily user feedback. 
Such numbers make it difficult for companies -- in particular with popular apps -- to employ a manual analysis on user feedback \cite{Guzman:Prio:2017}.
Therefore, gaining a deeper understanding of how to 1) filter noise and 2) how to extract requirements relevant information from user feedback is of high importance \cite{Maalej:REJ:2016}.
Recent advances in technology and scientific work enable new ways to tackle these challenges.
\section{Related Work} \label{sec:related_work}
In the paper ``Toward Data-Driven Requirements Engineering'', Maalej et al. \cite{Maalej:Toward:2016} describe the concept of \emph{User Feedback Analytics} which contains the two sub-categories \emph{Implicit Feedback} and \emph{Explicit Feedback}.
While \emph{Implicit Feedback} deals with usage data such as click events that are collected via software sensors on, e.g., a mobile device, \emph{Explicit Feedback} is concerned with written text such as app reviews.
We focus on \emph{Explicit Feedback}, which in the field of requirements engineering often includes either \emph{app reviews} \cite{Harman:App:2012, Guzman:Sentiment:2014, Maalej:REJ:2016}, \emph{tweets} \cite{ Guzman:Twitter:2017, Williams:Twitter:2017}, product reviews such as \emph{Amazon reviews} \cite{Kurtanovic:Mining:2017, kurtanovic:Rationale:2018}, a combination of reviews and product descriptions \cite{Johann:2017:app}, or a combination of platforms \cite{nayebi2017}.
User feedback is essential to practitioners, as it contains valuable insights such as bug reports and feature requests \cite{Pagano:App:2013}.
The classification of user feedback \cite{Maalej:REJ:2016} was a first step towards extracting such information.
Further studies \cite{Kurtanovic:Mining:2017, kurtanovic:Rationale:2018} looked at classified feedback to analyze and understand user rationale---the reasoning and justification of user decisions, opinions, and beliefs.
Once a company decides to integrate, for example, an innovative feature request in the software product, it will be forwarded to the release planning phase \cite{villarroel:release:2016}.
In this work, we focus on the classification of user feedback of app reviews and tweets.

\textbf{App Review Classification}.
Maalej et al. \cite{Maalej:REJ:2016} present experiments on classifying app reviews from the Google Play Store and the Apple AppStore using traditional machine learning.
In contrast to their work, we also apply deep learning, included tweets and work with two different languages (English and Italian).
Chen et al. \cite{chen2014ar} introduce \emph{AR-Miner}, a framework focusing on mining and ranking techniques to extract valuable information for developers following the idea of reducing manual effort.
Dhinakaran et al. \cite{Dhinakaran2018AppRA} perform app review classification and enhance existing approaches with active learning to reduce the annotation effort for experts. 

\textbf{Tweet Classification}.
Guzman et al. \cite{Guzman:Twitter:2017} and Williams and Mahmoud \cite{Williams:Twitter:2017} present studies that assess the technical value of tweets for software requirements. 
Williams and Mahmoud \cite{Williams:Twitter:2017} conclude that---after analyzing 4,000 tweets manually---about 51\% of the tweets contain technical information useful for requirements engineering. 
Similarly, Guzman et al. \cite{Guzman:Twitter:2017} show that about 42\% of their 1,350 manually analyzed tweets contain either bug reports, feature shortcomings, or feature requests.
Conceptually, both studies follow similar goals and structure by: first preprocessing the data; second classifying tweets into their specified categories; and third grouping similar tweets.
Guzman et al. \cite{Guzman:Twitter:2017} go one step further and present a weighted function to rank tweets by their relevance.
Compared to both papers, we have a strong focus on reporting feature engineering by testing diverse features and feature combinations (see Table \ref{tab:features}).
Further, we perform the classification on two different languages and employ deep learning as an addittonal experiment.
\section{Conclusion} \label{sec:conclusion}
In this study, we present a series of classification experiments to find requirements-relevant information in English app reviews as well as in English and Italian tweets.
We applied supervised machine learning and compared traditional machine learning and deep learning approaches.
We rely our results on a) 10,000 English and 15,000 Italian annotated tweets from telecommunication Twitter support accounts, and b) on 6,000 annotations of English app reviews.
Our results show that, within our setting, traditional machine learning can achieve comparable results to deep learning, although we collected thousands of annotations for each channel.
\section*{Acknowledgement}
The work presented in this paper was conducted within the scope of the Horizon 2020 project OpenReq, which is supported by the European Union under the Grant Nr. 732463.
The work was also supported by the “Forum4.0” project as part of the ahoi.digital funding line. 
We thank Davide Fucci for helping collect and analyze the Italian tweets.

\bibliographystyle{abbrv}
\bibliography{lib}
\end{document}